\documentclass[sigconf]{acmart}

\usepackage[inline]{enumitem}
\AtBeginDocument{%
  \providecommand\BibTeX{{%
    \normalfont B\kern-0.5em{\scshape i\kern-0.25em b}\kern-0.8em\TeX}}}

\acmConference{The 2nd Workshop on NLG for HRI}{December 18}{2020}

\setcopyright{none}
\renewcommand\footnotetextcopyrightpermission[1]{}
\begin{document}

\title{Enabling Robots to Draw and Tell: \\Towards Visually Grounded Multimodal Description Generation}

\author{Ting Han}
\email{ting.han@aist.go.jp}
\affiliation{%
  \institution{ Artificial Intelligence Research Center \\
  AIST, Tokyo, Japan}
}

\author{Sina Zarrie{\ss}}
\email{sina.zarriess@uni-jena.de}
\affiliation{%
  \institution{Friedrich Schiller University Jena \\
  Jena, Germany}}
%


\begin{abstract}
Socially competent robots should be equipped with the ability to perceive the world that surrounds them and communicate about it in a human-like manner. Representative skills that exhibit such ability include generating image descriptions and visually grounded referring expressions. In the NLG community, these generation tasks are largely investigated in non-interactive and language-only settings. However, in face-to-face interaction, humans often deploy multiple modalities to communicate, forming seamless integration of natural language, hand gestures and other modalities like sketches. To enable robots to describe what they perceive with speech and sketches/gestures, we propose to model the task of generating natural language together with free-hand sketches/hand gestures to describe visual scenes and reallife objects, namely, visually-grounded multimodal description generation. In this paper, we discuss the challenges and evaluation metrics of the task, and how the task can benefit from progress recently made in the natural language processing and computer vision realms, where related topics such as visually grounded NLG, distributional semantics, and photo-based sketch generation have been extensively studied.
 \end{abstract}

\maketitle

\section{Introduction}
A key way for robots to exhibit social intelligence is to show the ability of communicating what they perceive in the surrounding world as humans do,  with both verbal and non-verbal means such as drawing a trajectory in shared space or on a piece of paper (i.e., gesturing and sketching) to illustrate  the contour of an object. Research in human-robot interaction has a long-standing interest in embodied multimodal interaction.  A lot of work has been done on studying the complex interplay between speech and gestures, as well as understanding and generating speech and symbolic gestures in interaction \cite{kopp2006towards,kopp2008multimodal,sowa2009computational,lucking2010bielefeld,kranstedt2005incremental,han2018placing,han2015building}. In comparison, much less work has focused on generating iconic gestures/sketches together with natural language to describe reallife objects, albeit such descriptions are multimodal in nature.

Generating visually grounded multimodal descriptions is a challenging task: \begin{enumerate*}[label=\arabic*)] \item It requires the generation of both natural language and gesture/sketch, each of which is a standalone challenging task. Given a photo, the NLG component generates a sequence of words to convey the content symbolically; The gesture/sketch generation component translates the content into a sequence of sketch strokes to communicate iconically. It decides what, where and in which order to draw. \item The two modalities are not independent, but bear close semantic and temporal relations \cite{mcneill1992hand}. Iconic gestures complement or supplement the semantic content of the accompanied speech through their formal relevance to referents in the speech, thus they do not exhibit meanings on their own. Instead, 
 their meanings are largely designated by the accompanied speech, making multimodal alignment reasoning a critical component of the generation task. \item Moreover, in situated interaction, the generated descriptions unfold as timing sequences. The durations of a gesture/sketch and the accompanied utterance must fit with each other,  in order to be temporally aligned. Therefore, generating aligned descriptions requires reasoning of both the semantics and the timing of speech and gestures/sketches, making the generation task extremely challenging.\end{enumerate*}

\begin{figure}[!h]
\centering
\includegraphics[width=6.5cm]{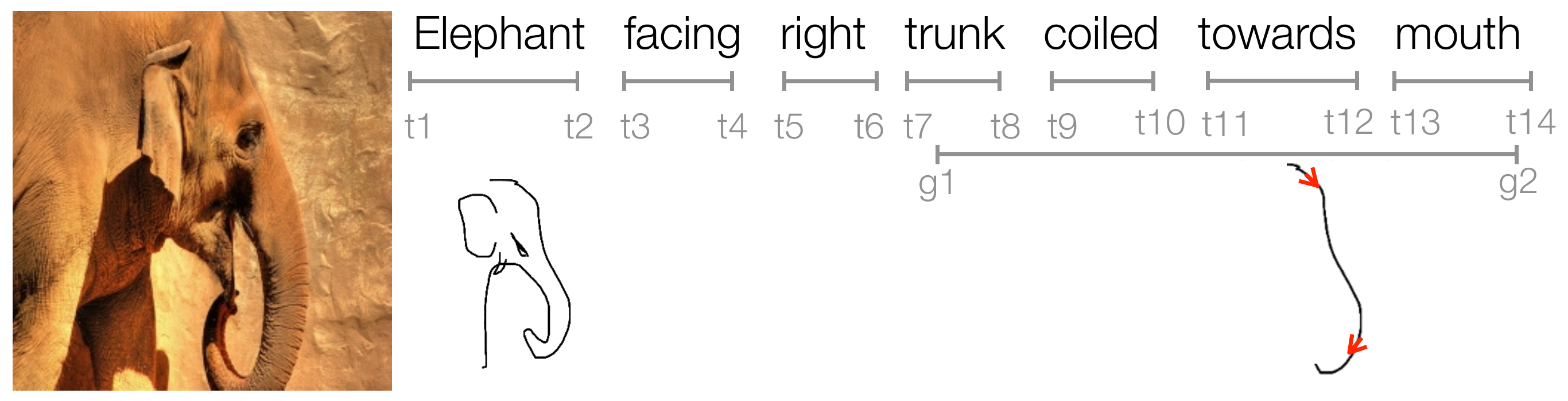}
\caption{A pseudo multimodal description of an elephant, composed of an utterance and a sketch stroke selected from a full sketch.  The red arrows indicate the sketch direction. 
\label{fig:example}}
\end{figure}

\vspace{-0.3cm}

For instance, in Figure~\ref{fig:example}, the utterance  ``\emph{elephant, facing right, trunk coiled towards mouth}''  and the sketch compose a multimodal description of the elephant in the photo. While the utterance describes the object symbolically, the sketch illustrates the shape of the trunk by visualizing its contour \cite{han2018learning,han2017natural}, which is difficult to convey symbolically with the word ``coiled''. Yet, the meaning of the sketch is ambiguous on its own.  It only becomes evident when jointly interpreting the utterance and the sketch, grounding the iconic concepts  encoded in both modalities to the same region of the photo (i.e., the trunk).

With the above observation in mind, we  propose to decompose the multimodal generation task into three sub-tasks: visually-grounded NLG, visually-grounded gesture/sketch generation, and multimodal alignment reasoning. The first two sub-tasks generate utterances and full sketches respectively. The multimodal alignment sub-task selects representative sketch strokes from full sketches,  infers optimal semantic and temporal relations between the utterances and the sketches, then outputs orchestrated descriptions. Finally, we convert the utterances to speech and pair the utterances with sketches displayed on a screen or projected to a gesture space together and duly to obtain aligned multimodal descriptions. 

Such a formulation will also benefit the proposed task from available pre-trained large scale language models,  sketch generation models, and large scale datasets, e.g.,  language and vision datasets such as MSCOCO and Visual Genome \cite{kazemzadeh2014referitgame,Lin2014}, and free-hand sketching dataset such as QuickDraw \cite{sangkloy2016sketchy}. To our knowledge, no large scale multimodal description datasets have been publicly available yet. Utilizing existing datasets is a good starting point for the proposed task.

\section{The Challenges}
\label{sec:challenges}

\subsection{Visually Grounded NLG}
The visually grounded NLG sub-task not only generates verbal object descriptions, it also provides important information for multimodal alignment. As aforementioned, sketch/gestures convey meaning through their formal relevance to referents in accompanied speech. Thus, we assume words and sketches that describe the same photo region are likely to align on semantic level. Grounding generated words to  visual inputs enables the identification of the region they describe.

Attention-based image caption models are a promising solution to our generation task. With an attention mechanism, image caption models focus on the relevant part of a given photo when generating each word. The ``focus'' is represented as a weight matrix: The region a word describes is with higher attention weights (\cite{xu2015show,Lu2017Adaptive}). With the attention matrix of each word,  we identify the region an utterance describes, then reason out sketch strokes that can be aligned with the utterance (see discussion in Section~\ref{subsec:alignment}).
 
\subsection{Photo-based Sketch Generation}
The sketch generation sub-task generates a sequence of  sketch strokes that exhibit contours of objects in a given photo (as shown in Figure~\ref{fig:example}), mimicking human sketching process by resembling various levels of sophistication and abstractness.

In computer vision, photo-based sketch generation is posited as a weakly supervised or unsupervised photo-to-sequence translation task  \cite{zhang2015end,xu2020deep,song2018learning,eitz2012humans}. Photo-sketch pairs are not strictly required for training sketch generation models. In recent works, neural network sketching models with encoding-decoding architecture typically learn a encoder network and a decoder network: the former encodes photos into feature vectors; the latter decodes  photo vectors to sketch sequences of target objects.


Note that  existing neural sketching models generate detailed sketches of objects, as they are trained on sketching datasets collected without timing constraints which might be too sophisticated to draw in situated interaction. In our task, we aim to accompany utterances with sketch strokes in human-robot interaction, where communication is often under timing pressure. Hence, a challenge to be addressed here is to constrain the sophistication of sketch strokes either during model training or via post-editing, so that they suit for situated communication.

\subsection{Composing Multimodal Descriptions}
\label{subsec:alignment}
To produce orchestrated descriptions,  the multimodal alignment sub-task first selects sketch strokes that represent salient iconic concepts of the target object, then determines which words are semantically and temporally  compatible with the selected strokes.  Several candidate descriptions might be initially proposed, from which the one that best suits the communicating purpose (e.g., emphasize a particular concept) is selected as the optimal description.

\textbf{Selecting representative sketch strokes.} Given a full sketch of a photo,  we select one or two sketch strokes to form an optimal multimodal description. The selection strategy here depends on the communication intention. For instance, to supplement a particular utterance, sketch strokes in the region the utterance describes should be selected; To be most informative, sketch strokes of the most salient part of the target object should be selected. 

\textbf{Multimodal alignment reasoning.} Regarding multimodal alignment, although word attention weights indicate which words correlate with selected strokes, note that not every word in natural language correlates with visual features in theory, hence, the attention weights only provide a very rough estimation of the alignment. To achieve more accurate alignment, we propose to reason the alignment according to semantics of words and sketches,  as accompanied speech and gestures bear close semantic relations \cite{han2017draw}.   Works in \emph{Distributional semantics} have shown that content from different modalities can be mapped to a joint embedding space, where vectors with  within shorted distances share similar semantic meaning \cite{bruni2014multimodal}. Clearly, iconic gestures and sketches are different, although they both convey iconicity. This approach could also be applied to measure semantic similarity of utterances and sketches. 


%




In a pilot study, we generated several multimodal descriptions with a simplified approach based on the above framework, and evaluated them under two settings:  1) display speech (converted from text with a TTS software) and sketches on a monitor, 2) enable a Nao robot to speak and gesture by projecting generated sketch strokes into Nao's gesture space. We found that, with right alignment, sketches/gestures are helpful in conveying iconicity.  Iconic gestures transformed from shorter and less sophisticated sketches are easier to interpret and more helpful. Generating iconic gestures based on sketches would be an interesting topic for future research.

 \vspace{-0.2cm}
\subsection{Evaluation}
 An informative evaluation should reflect the quality of individual modalities and how well the two modalities are aligned.  We propose to combine human evaluation with automatic evaluation to  assess the overall generation quality \cite{belz2006comparing}.

\textbf{Automatic evaluation} suits for fast evaluations during model development. We suggest to evaluate NLG quality with metrics in NLG such as BLEU\cite{papineni2002bleu,vedantam2015cider}, while evaluating  the informativeness of sketches and multimodal descriptions with sketch-based and multimodal image retrieving tasks: A higher retrieving accuracy indicates better description quality. Note that these automatic evaluations can not measure temporal alignment between modalities, therefore, we resort to human evaluation to assess multimodal alignment quality. \textbf{Human evaluation}  offers insights into multimodal alignment quality as miss-alignment impairs overall interpretation of the descriptions, especially the interpretation of sketches/gestures. Interactive referring games  between humans and robots that reveal how  humans understand the generated descriptions, are good testbeds for such evaluation \cite{kazemzadeh2014referitgame}.

\section{Conclusion and Future work}
We proposed the task of visually grounded multimodal description generation. Along with discussions on the challenges regarding NLG, free-hand sketch generation, multimodal alignment, and evaluation methods, we  pointed out how existing works of visually grounded NLG, photo-based sketch generation, distributional semantics, and their respective evaluation metrics benefit the proposed task.   We believe, solving the task will lead to more natural multimodal human-robot interaction in scenarios where robots can communicate what they perceive in the world that surrounds them.

\clearpage
\section*{Acknowledgments}
 This work is supported by the New Energy and Industrial
Technology Development Organization (NEDO) of Japan.


\bibliographystyle{ACM-Reference-Format}
\bibliography{refs}


\end{document}